\documentclass[letterpaper]{article} 
\usepackage[preprint]{aaai2027}  
\usepackage[hyphens]{url}  
\usepackage{graphicx} 
\usepackage{natbib}  
\usepackage{caption} 
\usepackage{algorithm}
\usepackage{algorithmic}

\usepackage{newfloat}
\usepackage{listings}
\DeclareCaptionStyle{ruled}{labelfont=normalfont,labelsep=colon,strut=off} 
\floatstyle{ruled}
\newfloat{listing}{tb}{lst}{}
\floatname{listing}{Listing}

\usepackage{booktabs}

\usepackage{amsmath}
\usepackage{amssymb}

\title{MetaRank: Task-Aware Metric Selection for Model Transferability Estimation}
\author {
    Yuhang Liu\textsuperscript{\rm 1},
    Wenjie Zhao\textsuperscript{\rm 2},
    Xin Wang\textsuperscript{\rm 1},
    Yunhui Guo\textsuperscript{\rm 2}
}
\affiliations {
    \textsuperscript{\rm 1}Fudan University\\
    \textsuperscript{\rm 2}The University of Texas at Dallas\\
    liuyuhang25@m.fudan.edu.cn, wxz220013@utdallas.edu,
    xwang11@fudan.edu.cn, yunhui.guo@utdallas.edu
}

\begin{document}

\maketitle


\begin{abstract}
Selecting an appropriate pre-trained source model is a critical, yet computationally expensive, task in transfer learning. Model Transferability Estimation (MTE) methods address this by providing efficient proxy metrics to rank models without full fine-tuning. In practice, the choice of which MTE metric to use is often ad hoc or guided simply by a metric's average historical performance. However, we observe that the effectiveness of MTE metrics is highly task-dependent and no single metric is universally optimal across all target datasets. To address this gap, we introduce MetaRank, a meta-learning framework for automatic, task-aware MTE metric selection. MetaRank adopts a retrieve-and-rerank cascade. A lightweight retrieval stage first narrows the metric pool using performance observed on similar meta-training datasets. To address the heterogeneous definitions and scales of MTE scores, the reranking stage represents each retrieved metric through the pairwise source-model ordering it induces on the target dataset. A product-kernel regressor combines dataset and ordering similarities to refine the ranking, enabling cross-metric transfer from historical metrics with similar ordering behavior. MetaRank then ranks the retrieved metrics and selects the most appropriate one to guide source-model selection on an unseen target dataset. Extensive experiments across 11 pre-trained models, 13 candidate MTE metrics, and 14 target datasets demonstrate that MetaRank significantly outperforms all compared baselines.
\end{abstract}


\section{Introduction}
\label{sec:intro}
Transfer learning leverages knowledge from a source domain to improve learning on a target domain while reducing annotation and computational costs \cite{pan2009survey,zhuang2020comprehensive}. The dominant transfer learning paradigm with deep neural networks, which involves pre-training on a source domain and subsequently fine-tuning on a target domain, has proven effective across diverse application areas \cite{kornblith2019better, hu2019strategies}.
\begin{figure}[t]
  \centering
   \includegraphics[width=\linewidth]{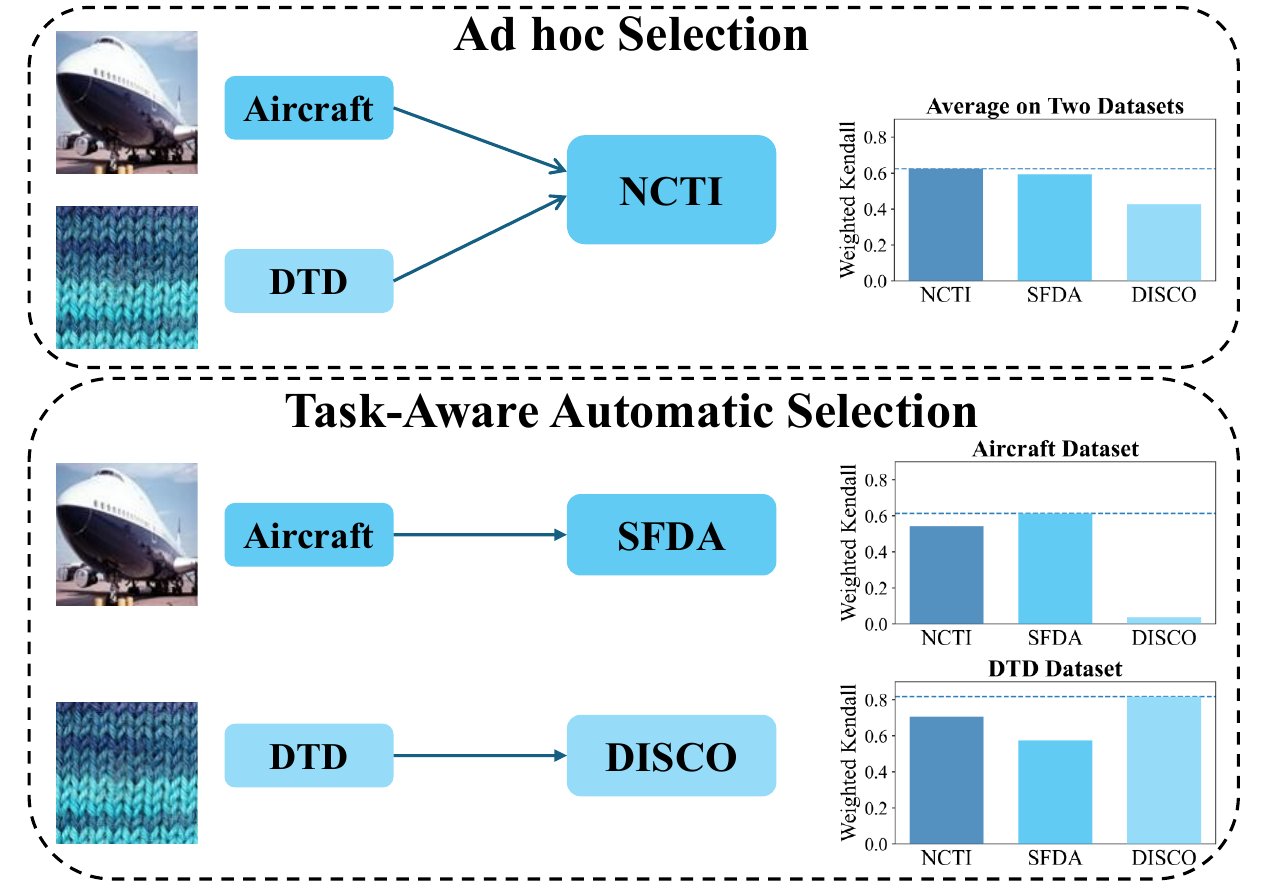}
   \caption{Comparison of ad hoc and task-aware MTE metric selection. Ad hoc selection employs a single MTE metric (e.g., NCTI), which may achieve the highest average performance but fails to identify the best-performing metric for each target dataset. 
In contrast, task-aware selection adapts the choice of metric for each dataset (e.g., SFDA for Aircraft and DISCO for DTD).}
   \label{fig:intro}
\end{figure}
However, transfer effectiveness varies markedly depending on the choice of pre-trained source model, as different models can yield substantially different performance on the same target dataset. Appropriate source models can improve downstream performance, while inappropriate ones may lead to negative transfer \cite{pan2009survey}. Therefore, selecting the most appropriate source model is a critical challenge, especially given the rapidly expanding pool of publicly available pre-trained models. A straightforward approach is to fine-tune each candidate source model on the target dataset to identify the best one. However, such a brute-force assessment is computationally infeasible, motivating the adoption of more efficient model transferability estimation (MTE).

MTE aims to use computationally efficient transferability scores to estimate a source model's downstream performance with reduced overhead. This goal has driven the development of various metrics that quantify transferability through different analytical approaches \cite{ding2024model, guo2024kite}. For example, LEEP \cite{nguyen2020leep} estimates the empirical conditional distribution from the joint distribution induced by applying the pre-trained model to target data. In practice, practitioners often select a fixed MTE metric based on its average performance over historical datasets and apply it to new target datasets. However, we observed that although a particular MTE metric may achieve the highest average performance, it does not necessarily select the best source model for every target dataset. As shown in Fig.~\ref{fig:intro}, while NCTI \cite{wang2023NCTI} achieves the highest average performance across the two target datasets, this overall success often masks dataset-dependent variability. For example, SFDA \cite{shao2022SFDA} performs best on Aircraft \cite{maji2013aircraft}, whereas DISCO \cite{zhang2025disco} performs best on DTD \cite{cimpoi2014dtd}.
These observations highlight the importance of automatically selecting the best-performing MTE metric for each target dataset to achieve superior performance. This finding is conceptually aligned with the no-free-lunch
principle \cite{freelunch}, which suggests that no single method can be optimal across all problem settings. Therefore, effective MTE metric selection is critical; however, existing approaches remain largely ad hoc and often result in suboptimal outcomes.

In this paper, we introduce MetaRank, a meta-learning framework for automatic, task-aware MTE metric selection. MetaRank combines historical evidence from meta-training datasets with the source-model ordering behavior of candidate metrics on the target dataset. Specifically, it adopts a retrieve-and-rerank cascade. The retrieval stage identifies a compact set of promising metrics based on their historical performance on meta-training datasets with similar meta-features. Beyond reducing online metric-evaluation cost, this stage prunes less promising candidates and provides the reranker with a more relevant search space. The reranking stage then converts the scores of each retrieved metric into pairwise source-model preferences, providing a shared representation for metrics with heterogeneous definitions and scales. It further employs product-kernel ridge regression to predict metric utility by jointly modeling similarities in dataset meta-features and ordering behaviors, enabling fine-grained metric ranking.

The main contributions of this paper are summarized as follows:
\begin{itemize}
\item We show that the effectiveness of MTE metrics is highly task-dependent, which motivates the development of an automatic task-aware selection method.

\item We propose MetaRank, a meta-learning framework with a retrieve-and-rerank architecture. MetaRank maps heterogeneous metrics into a shared behavioral space via pairwise-order representations, and employs a product-kernel to jointly model dataset meta-features and ordering behavior for task-aware metric selection.

\item Extensive experiments across 11 pre-trained models, 13 candidate MTE metrics, and 14 target datasets demonstrate that MetaRank significantly outperforms all compared baselines. Moreover, MetaRank exhibits robust generalization, successfully adapting to unseen metrics without retraining.

\end{itemize}

\section{Related Work}
\subsection{Model Transferability Estimation}
Model transferability estimation (MTE) aims to predict how well a pre-trained model will perform on a target task after fine-tuning, without incurring the cost of full fine-tuning. MTE metrics are commonly categorized into static and dynamic approaches \cite{ding2024model}. Static approaches obtain transferability scores directly from statistical properties of features or logits produced by candidate source models on the target dataset. They can be further divided into four categories: feature-structure-based methods, such as GBC \cite{pandy2022GBC} and NCTI \cite{wang2023NCTI}; Bayesian-statistics-based methods, such as LEEP \cite{nguyen2020leep} and its variants \cite{li2021nleep}; information-theoretic methods, such as negative conditional entropy (NCE) \cite{tran2019NCE}; and matrix-analysis-based methods, such as DISCO \cite{zhang2025disco}, which weights task-specific component scores by singular-value ratios. Dynamic approaches, by contrast, map the original static information into a different space via mapping functions or learning frameworks before computing scores. Representative methods include SFDA~\cite{shao2022SFDA}, which projects target features into a self-challenging Fisher space, SA~\cite{khoba2025SA}, which applies spread-and-attract perturbations to feature embeddings, and ETran~\cite{gholami2023etran}, an energy-based method. Despite the wide variety of MTE metrics, their diverse underlying mechanisms lead to substantial performance variability across datasets, motivating a task-aware automatic metric selection method.

\subsection{Meta-Learning}
Meta-learning transfers knowledge across meta-training tasks to support predictions or decisions on an unseen task \cite{hospedales2021metasurvey,vettoruzzo2024advances}. In automated machine learning, tasks are commonly characterized by dataset meta-features, which, together with historical algorithm performance, guide the recommendation of algorithms or configurations for a new dataset \cite{qin2024metaood}. Such meta-features may be manually designed or learned directly from data. Task2Vec \cite{achille2019task2vec} and Dataset2Vec \cite{jomaa2021dataset2vec} are representative methods for learning dataset meta-features, using Fisher-information-based task embeddings and direct dataset-level encoding, respectively. Within MTE, methods such as SynLearn \cite{ding2022synlearn} and ModelSpider \cite{zhang2023modelspider} learn representations of pre-trained models and target tasks and use their interactions to estimate model--task compatibility, thereby ranking source models. In contrast, our problem is to estimate the compatibility between target datasets and MTE metrics, thereby ranking the metrics themselves.

\section{Preliminaries}
\label{sec:preliminary}
\subsection{Model Transferability Estimation}
\label{sec:mte}
In model transferability estimation (MTE), we are given a model zoo $\mathcal{M}=\{m_1,\dots,m_M\}$ and a dataset $\mathcal{D}_d=\{(x_j,y_j)\}_{j=1}^{n_d}$. For each source model $m_i\in\mathcal{M}$, let $m_i^*$ denote the model obtained after full fine-tuning on $\mathcal{D}_d$. Its ground-truth transferability is defined as $T_d(m_i)=\mathcal{E}(m_i^*)$, where $\mathcal{E}$ denotes a downstream evaluation metric such as accuracy or F1-score. Since obtaining $T_d(m_i)$ for all source models requires full fine-tuning, an MTE metric $q$ instead computes an efficient transferability score $S_{d,q}(m_i)$ for each source model, yielding the score vector
\(\mathbf{s}_{d,q}
=[S_{d,q}(m_1),\ldots,S_{d,q}(m_M)]^\top\). We define the utility of $q$ on $\mathcal{D}_d$ as the rank agreement between the estimated scores $\{S_{d,q}(m_i)\}_{i=1}^{M}$ and the ground-truth downstream performances $\{T_d(m_i)\}_{i=1}^{M}$. We measure this utility using the weighted Kendall rank correlation coefficient $\tau_w$ \cite{vigna2015kendall}. Let $S_i=S_{d,q}(m_i)$ and $T_i=T_d(m_i)$. The coefficient is defined as
\begin{equation}
\resizebox{0.9\columnwidth}{!}{$\displaystyle
\tau_w =
\frac{
\sum_{i<j} w_{ij}\sigma(S_i-S_j)\sigma(T_i-T_j)
}{
\sqrt{
\left(\sum_{i<j} w_{ij}\sigma^2(S_i-S_j)\right)
\left(\sum_{i<j} w_{ij}\sigma^2(T_i-T_j)\right)
}
}
$}
\label{eq:weighted-kendall}
\end{equation}
where $\sigma$ is the sign function and $w_{ij}$ denotes the weight associated with the comparison between source models $m_i$ and $m_j$. Accordingly, the utility of metric $q$ on dataset $\mathcal{D}_d$ is defined as $u_{d,q}=\tau_w\!\left(\{S_{d,q}(m_i)\}_{i=1}^{M},\{T_d(m_i)\}_{i=1}^{M}\right).$
A larger $u_{d,q}$ indicates that metric $q$ more accurately recovers the ground-truth source-model ranking on $\mathcal{D}_d$.

\subsection{Necessity of MTE Metric Selection}
\begin{figure}[t]
  \centering
   \includegraphics[width=\linewidth]{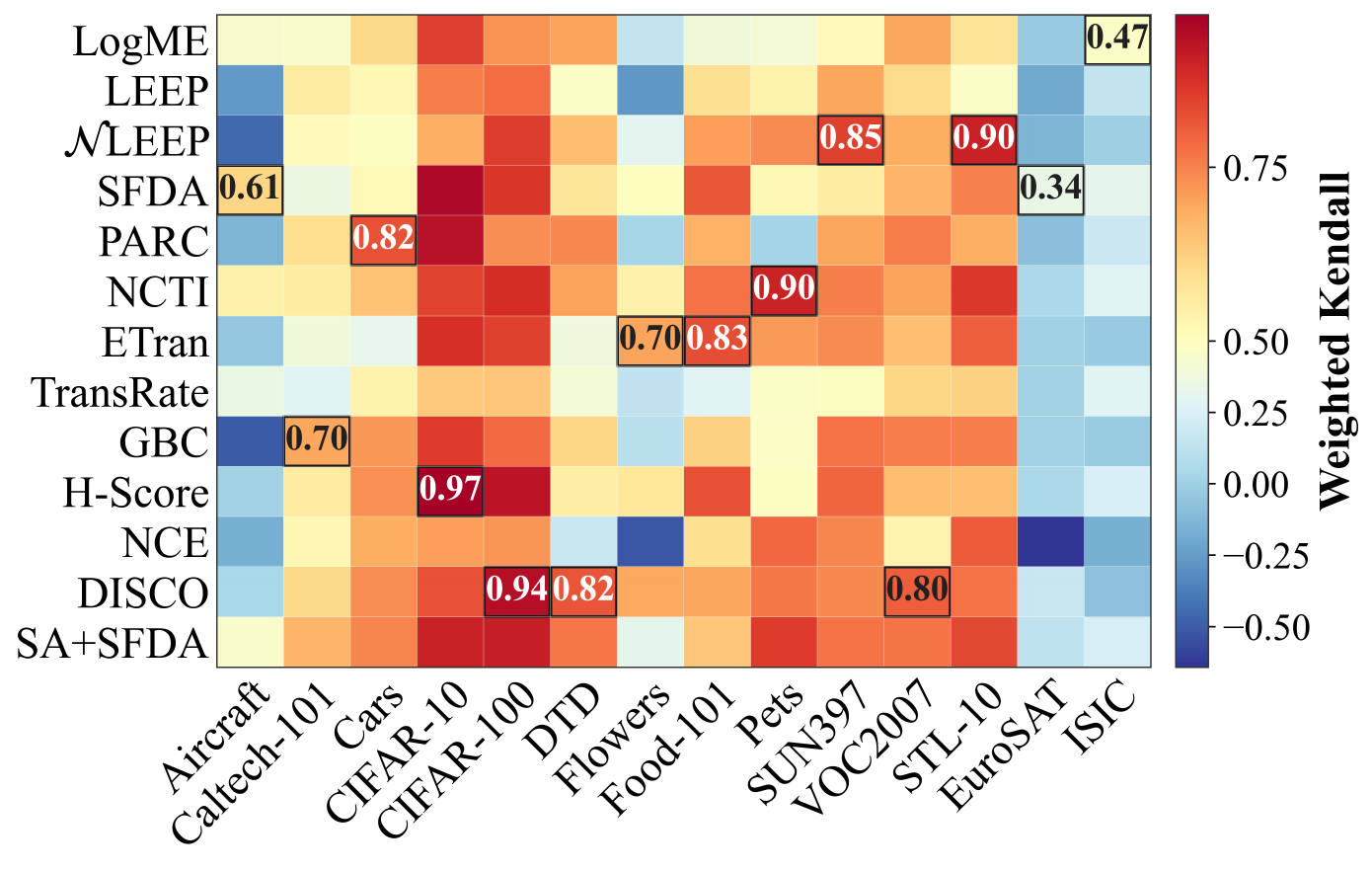}
   \caption{Weighted Kendall's $\tau_w$ of MTE metrics across target datasets. Black outlines mark the best metric for each dataset. The results demonstrate that no single metric is universally optimal.}
   \label{fig:heatmap}
\end{figure}

MTE metrics serve as a critical tool for guiding the selection of the most appropriate pre-trained source models for various target datasets. However, as shown in Fig.~\ref{fig:heatmap}, no single MTE metric consistently achieves high performance across all target datasets, and metric utility exhibits substantial dataset-dependent variability. For example, H-score \cite{bao2019HScore} achieves $\tau_w=0.97$ on CIFAR-10 \cite{krizhevsky2009cifar}, but only $\tau_w=0.01$ on Aircraft \cite{maji2013aircraft}.

This variability arises because different metrics rely on mechanisms that respond differently to dataset properties such as sample size, class granularity, and feature distribution.
H-score measures class separability using the overall feature covariance and the covariance of class-conditional means \cite{bao2019HScore}. CIFAR-10 contains a large number of samples per class and comprises relatively coarse-grained categories, providing comparatively stable estimates of these statistics \cite{krizhevsky2009cifar}. In contrast, Aircraft contains far fewer samples per class and demands fine-grained discrimination among visually similar categories \cite{maji2013aircraft}. These conditions make class means and covariance structures harder to estimate accurately, reducing H-score's ability to distinguish candidate models. This interpretation aligns with empirical evidence that H-score becomes unstable when sample size is small relative to feature dimensionality and less reliable on fine-grained datasets \cite{ibrahim2022newer,abou2024onesize}.
Such dataset-dependent variability indicates that selecting an MTE metric without considering the target dataset can yield inaccurate source-model rankings.

To the best of our knowledge, little prior work has directly addressed MTE metric selection. A recent study provides a decision tree that recommends transferability metrics according to dataset properties and computational constraints \cite{abou2024onesize}. For example, if the target dataset is small-scale and fine-grained, the decision tree recommends SFDA \cite{shao2022SFDA} or LogME \cite{you2021logme}; otherwise, it applies additional manually specified criteria. However, these rules are manually derived from empirical observations rather than learned from data, and their effectiveness is not evaluated on held-out target datasets. Moreover, the rules are tied to a fixed set of evaluated metrics, limiting their extensibility to newly developed metrics. These limitations underscore the need for an automated, generalizable framework capable of selecting appropriate MTE metrics for previously unseen target datasets.

\section{Proposed Method}
\label{sec:method}
Given a pool of MTE metrics $\mathcal{Q}=\{q_1,\ldots,q_N\}$, MetaRank formulates metric selection as a task-level meta-learning problem over a meta-training set $\mathcal{D}_{\mathrm{meta}}=\{\mathcal{D}_1,\ldots,\mathcal{D}_L\}$. Each dataset $\mathcal{D}_d \in \mathcal{D}_{\mathrm{meta}}$ represents a meta-training task, providing the dataset meta-feature vector $\mathbf{c}_d$, the metric score vectors $\{\mathbf{s}_{d,q}\}_{q=1}^{N}$, and the metric utilities $\{u_{d,q}\}_{q=1}^{N}$. The meta-feature $\mathbf{c}_d$ is constructed from frozen CLIP ViT-B/32 representations \cite{clip} and basic dataset statistics, such as class distribution, with details in the Supplementary Material. By capturing dataset--metric relationships across these tasks during offline meta-training, MetaRank transfers the acquired knowledge to efficiently rank metrics for an unseen target dataset during online inference. 

\begin{figure*}[t]
  \centering
   \includegraphics[width=0.9\linewidth]{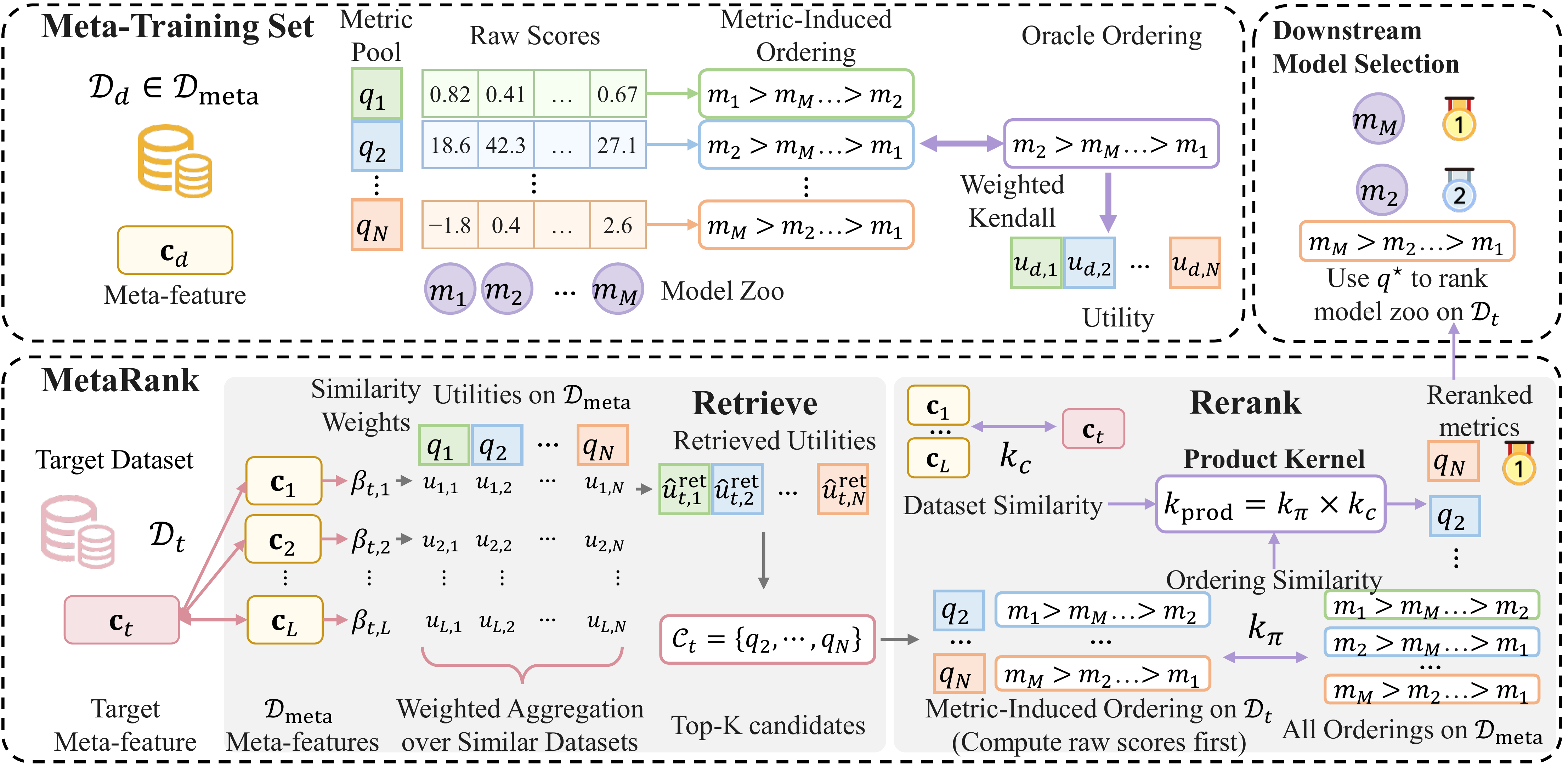}
   \caption{Overview of MetaRank. During meta-training, each dataset is represented by meta-features and paired with metric-induced source-model orderings and observed utilities. For a target dataset, similarity-guided retrieval selects the top-\(K\) candidate metrics, while product-kernel reranking combines meta-feature similarity with source-model ordering behavior to identify the most appropriate metric for downstream source-model selection.}
   \label{fig:workflow}
\end{figure*}

As shown in Fig.~\ref{fig:workflow}, MetaRank adopts a retrieve-and-rerank cascade consisting of similarity-guided retrieval and fine-grained product-kernel reranking. The retrieval stage estimates coarse metric utilities by transferring observed utility patterns from similar meta-training datasets. The reranking stage further encodes the ordering behavior induced by each retrieved metric in a shared pairwise-order space and couples it with dataset meta-features through a product kernel, enabling fine-grained utility prediction and cross-metric transfer. The two stages are detailed in the following subsections.

\subsection{Similarity-Guided Retrieval}
\label{sec:retrieval}
The retrieval stage serves as a lightweight filter that narrows the candidate set, thereby reducing the computational cost of fine-grained reranking. A metric that performs well on meta-training datasets similar to the target dataset is expected to be a promising candidate. Let $\widetilde{\mathbf{c}}_d$ denote the meta-feature vector of meta-training dataset $\mathcal{D}_d$, standardized using statistics estimated from the meta-training set and subsequently $\ell_2$-normalized. We measure the similarity between two datasets via an exponential kernel over their cosine distance:
\begin{equation}
k_c(d,d')
=
\exp\!\left[
-\frac{
1-\widetilde{\mathbf{c}}_d^\top\widetilde{\mathbf{c}}_{d'}
}{
h_c
}
\right],
\label{eq:dataset-similarity}
\end{equation}
where $h_c$ is set to the median positive pairwise cosine distance among the meta-training datasets.

During online inference, the dataset meta-feature vector $\mathbf{c}_t$ is first constructed for the unseen target dataset $\mathcal{D}_t$ following the same procedure. It is then transformed using the meta-training preprocessing statistics, yielding $\widetilde{\mathbf{c}}_t$. We estimate the coarse utility of each metric $q\in\mathcal{Q}$ through similarity-weighted aggregation:
\begin{equation}
\beta_{t,d}
=
\frac{
k_c(t,d)
}{
\sum_{j=1}^{L}k_c(t,j)
},
\qquad
\widehat{u}^{\mathrm{ret}}_{t,q}
=
\sum_{d=1}^{L}
\beta_{t,d}u_{d,q}.
\label{eq:retrieval-utility}
\end{equation}
Thus, meta-training datasets that are more similar to the target dataset contribute more to its coarse metric-utility estimates. The $K$ metrics from the full metric pool $\mathcal{Q}$ with the highest estimated utilities form the retrieved candidate set $\mathcal{C}_t$, where $K$ is the retrieval budget. This provides the fine-grained reranking stage with a compact and relevant candidate pool, reducing interference from unlikely metrics while avoiding unnecessary MTE score computation.

\subsection{Product-Kernel Reranking}
\label{sec:reranking}
Given the retrieved candidate set $\mathcal{C}_t$, the reranking stage further differentiates among the candidate metrics by exploiting the ordering behaviors induced by their score vectors. Its key idea is to represent heterogeneous MTE metrics through the source-model orderings they induce, providing a shared pairwise-order space in which utility evidence can be transferred across both datasets and metrics. A product kernel then couples this ordering behavior with dataset meta-features to predict fine-grained metric utilities.

Although different MTE metrics produce scores with heterogeneous definitions and scales, they evaluate the same source-model pool. The fixed source-model pairs therefore provide a shared reference for representing their ordering behaviors. Specifically, for metric $q$ on meta-training dataset $\mathcal{D}_d$, we encode the relative ordering of all source-model pairs:
\begin{equation}
\mathbf{z}_{d,q}
=
\left[
\sigma\!\left(
S_{d,q}(m_i)-S_{d,q}(m_j)
\right)
\right]_{1\leq i<j\leq M}
\in\{-1,0,1\}^{P},
\label{eq:pairwise-comparison}
\end{equation}
where $P=M(M-1)/2$. The resulting pairwise-sign vector is normalized to form the pairwise-order representation
$\boldsymbol{\phi}_{d,q}
=
\mathbf{z}_{d,q}/\max(\|\mathbf{z}_{d,q}\|_2,\epsilon)$,
where $\epsilon>0$ prevents division by zero.
Each dimension records the preference induced by metric $q$ for a fixed source-model pair while discarding metric-specific score magnitudes. Since its coordinates are defined by the shared source-model pairs, $\boldsymbol{\phi}_{d,q}$ provides a unified representation of ordering behavior that can be directly compared across datasets and metrics. Based on these representations, we define the pairwise-order kernel
\begin{equation}
k_\pi\!\left(
\boldsymbol{\phi}_{d,q},
\boldsymbol{\phi}_{d',q'}
\right)
=
\exp\!\left[
-\frac{
1-
\boldsymbol{\phi}_{d,q}^{\top}
\boldsymbol{\phi}_{d',q'}
}{
h_\pi
}
\right],
\label{eq:pairwise-order-kernel}
\end{equation}
where $h_\pi$ is set to the median positive pairwise-order distance among all meta-training pairs.

To condition utility transfer jointly on dataset similarity and ordering similarity, we define a product kernel over dataset--metric pairs:
\begin{equation}
k_{\mathrm{prod}}\!\left(
(d,q),(d',q')
\right)
=
k_c(d,d')\,
k_\pi\!\left(
\boldsymbol{\phi}_{d,q},
\boldsymbol{\phi}_{d',q'}
\right),
\label{eq:product-kernel}
\end{equation}
where $k_c$ is the dataset similarity kernel defined in Eq.~\eqref{eq:dataset-similarity}. The product kernel assigns high similarity only when both the datasets and the metric-induced orderings are similar. It therefore allows utility evidence to be transferred across different metrics when they exhibit similar ordering behaviors on similar datasets. 

During offline meta-training, we employ kernel ridge regression (KRR) with the product kernel to learn the mapping from meta-training dataset--metric pairs to their utilities. The KRR coefficient vector is obtained in closed form as
\begin{equation}
\begin{gathered}
\boldsymbol{\alpha}
=
\left(
\mathbf{K}_{\mathrm{prod}}
+
\lambda\mathbf{I}
\right)^{-1}
\mathbf{u},
\qquad
\lambda=1/L,\\
\left[
\mathbf{K}_{\mathrm{prod}}
\right]_{(d,q),(d',q')}
=
k_{\mathrm{prod}}\!\left(
(d,q),(d',q')
\right),
\end{gathered}
\label{eq:product-krr-fit}
\end{equation}
where $\mathbf{K}_{\mathrm{prod}}$ denotes the corresponding Gram matrix and $\mathbf{u}=
[u_{1,1},\ldots,u_{1,N},
\ldots,
u_{L,1},\ldots,u_{L,N}]^\top$.

During online inference, the target meta-feature representation $\widetilde{\mathbf{c}}_t$ obtained in the retrieval stage is reused. Only the retrieved metrics $q\in\mathcal{C}_t$ are evaluated on the unseen target dataset $\mathcal{D}_t$, yielding $K$ target score vectors $\{\mathbf{s}_{t,q}\}_{q\in\mathcal{C}_t}$. Each score vector is transformed into the pairwise-order representation $\boldsymbol{\phi}_{t,q}$ using the same pairwise sign encoding and normalization procedure. The utility of each retrieved metric is then predicted through the kernel expansion
\begin{equation}
\widehat{u}_{t,q}
=
\sum_{d=1}^{L}
\sum_{q'=1}^{N}
\alpha_{d,q'}
k_c(t,d)
k_\pi\!\left(
\boldsymbol{\phi}_{t,q},
\boldsymbol{\phi}_{d,q'}
\right),
\label{eq:utility-prediction}
\end{equation}
where $\alpha_{d,q'}$ is the coefficient learned during offline meta-training for the meta-training dataset--metric pair $(d,q')$. The kernel expansion aggregates evidence from all such pairs, with each contribution jointly determined by the learned coefficient, dataset similarity, and ordering similarity. Since the summation spans all metrics, it enables cross-metric transfer from historical pairs involving metrics $q'\neq q$ with similar ordering behavior on related datasets. Moreover, the influence of each meta-training dataset--metric pair can be examined through its individual term in the expansion.

Finally, the retrieved metrics are sorted in descending order of their predicted utilities, and the final recommended metric is
$
\widehat{q}_t
=
\arg\max_{q\in\mathcal{C}_t}
\widehat{u}_{t,q}.
$
When a complete ranking over $\mathcal{Q}$ is required, the reranked metrics in $\mathcal{C}_t$ are placed first according to $\widehat{u}_{t,q}$, while the remaining metrics are appended in descending order of their retrieval-stage estimates $\widehat{u}^{\mathrm{ret}}_{t,q}$.

\section{Experiments}
\subsection{Experimental Setup}
\noindent \textbf{Datasets and Pre-trained Models.} We adopt 14 widely used image-classification datasets, including FGVC-Aircraft \cite{maji2013aircraft}, Caltech-101 \cite{fei2004caltech101}, Stanford Cars \cite{krause2013cars}, CIFAR-10 and CIFAR-100 \cite{krizhevsky2009cifar}, DTD \cite{cimpoi2014dtd}, Oxford-102 Flowers \cite{nilsback2008flower}, Food-101 \cite{bossard2014food}, Oxford-IIIT Pets \cite{parkhi2012pets}, SUN397 \cite{xiao2010sun}, VOC2007 \cite{everingham2010voc}, STL-10 \cite{coates2011stl10}, EuroSAT \cite{helber2019eurosat}, ISIC \cite{gutman2016ISIC}. The corresponding fine-tuned accuracies are obtained from \cite{shao2022SFDA}. We construct a pool of 11 widely used ImageNet-pre-trained models including ResNet-34/50/101/152 \cite{he2016res}; DenseNet-121/169/201 \cite{huang2017densely}; MnasNet-A1 \cite{tan2019mnasnet}; MobileNetV2 \cite{sandler2018mobilenetv2}; GoogLeNet \cite{szegedy2015google}; and Inception v3 \cite{szegedy2016inception}.

\noindent \textbf{Baselines.}
We consider 13 MTE metrics: LogME \cite{you2021logme}, LEEP \cite{nguyen2020leep}, $\mathcal{N}$LEEP \cite{li2021nleep}, SFDA \cite{shao2022SFDA}, PARC \cite{parc}, NCTI \cite{wang2023NCTI}, ETran \cite{gholami2023etran}, TransRate \cite{transrate}, GBC \cite{pandy2022GBC}, H-Score \cite{bao2019HScore}, NCE \cite{tran2019NCE}, DISCO \cite{zhang2025disco}, and SA+SFDA \cite{khoba2025SA}. We compare MetaRank with two groups of baselines. \textbf{(i) Fixed and random selection.}
Each \textbf{fixed-metric} baseline applies one metric to every target dataset, while \textbf{Random} uniformly samples a random permutation and selects its top-ranked metric. \textbf{(ii) Meta-learners.}
\textbf{Global Best (GB)} ranks candidate metrics by their mean utility on the meta-training datasets and selects the top-ranked metric. \textbf{ARGOSMART (AS)} \cite{nikolic2013as} transfers the best metric from the nearest training dataset, whereas \textbf{ISAC} \cite{kadioglu2010isac} recommends the best metric within the nearest dataset cluster. \textbf{ALORS} \cite{misir2017alors} combines matrix factorization with dataset features, and \textbf{NCF} \cite{he2017ncf} replaces its linear interaction with a neural predictor. All feature-based baselines use the same dataset meta-features as MetaRank; NCF represents the known candidate metrics using one-hot identities.

\noindent \textbf{Implementation Details.} We adopt leave-one-dataset-out (LODO) evaluation. In each of the 14 outer folds, one dataset is held out for testing, while the remaining $L=13$ datasets form the training set. The retrieval budget is set to $K=5$. All preprocessing statistics and kernel bandwidths are estimated independently within each fold. Each dataset meta-feature is constructed from 2,048 randomly sampled instances using seed 42.

\noindent \textbf{Evaluation.} We evaluate using the selected-metric rank, i.e., the ground-truth rank of the selected metric among the \(N=13\) candidates, where lower is better. We also report the \(\tau_w\) of the selected metric and Spearman correlation between the predicted and ground-truth rankings of candidate metrics. Statistical significance is assessed using two-sided paired Wilcoxon signed-rank tests on the \(\tau_w\) values of the metrics selected by each method with Holm correction.

\begin{figure*}[ht]
  \centering
   \includegraphics[width=\linewidth]{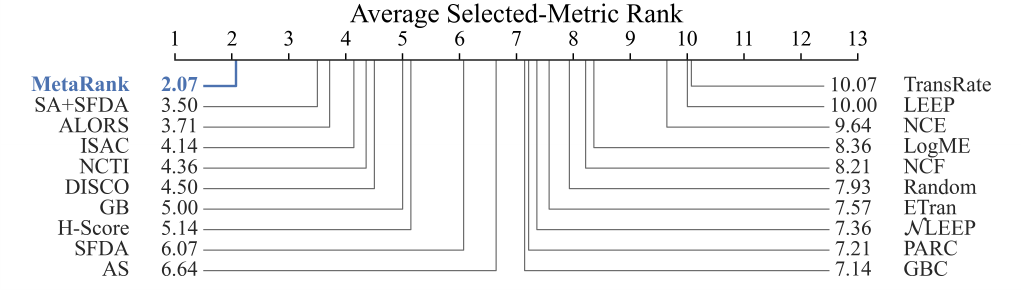}
   \caption{Average selected-metric rank. Methods are plotted according to the ground-truth rank of their selected metrics, averaged over all target datasets (lower is better). MetaRank achieves the lowest rank, outperforming all baselines.}
   \label{fig:main-rank}
\end{figure*}

\subsection{Main Results}
\begin{figure}[t]
  \centering
   \includegraphics[width=\linewidth]{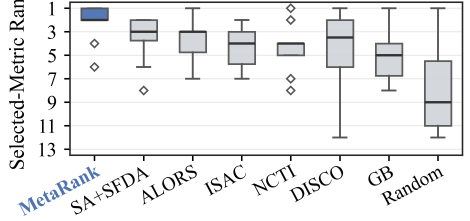}
   \caption{Distribution of selected-metric ranks. MetaRank concentrates most selections near the optimum and achieves the lowest median rank.}
   \label{fig:main-box}
\end{figure}

Across the 14 target datasets, MetaRank achieves the strongest overall metric-selection performance. As shown in Fig.~\ref{fig:main-rank}, it obtains the lowest average selected-metric rank of 2.07, outperforming SA+SFDA, the best fixed-metric baseline, by 1.43 ranking positions and ALORS, the best competing meta-learner, by 1.64 positions. Table~\ref{tab:main} provides consistent evidence in terms of selected $\tau_w$, for which MetaRank achieves the highest value of 0.7062. Furthermore, its gains in the selected $\tau_w$ relative to each baseline are statistically significant under two-sided paired Wilcoxon signed-rank tests with Holm correction, with all adjusted $p$-values below 0.05. Among the compared baselines that produce complete candidate-metric rankings, MetaRank also achieves the highest Spearman correlation of 0.5183. The per-dataset distributions in Fig.~\ref{fig:main-box} further show that MetaRank's advantage extends beyond average performance. MetaRank achieves a median rank of 2 and an interquartile range (IQR) of 1--2. It selects a top-two metric on 12 of the 14 target datasets, including the ground-truth best metric on 5 datasets. Its worst observed rank is 6, whereas every baseline drops to rank 7 or worse on at least one dataset. MetaRank therefore provides both a tighter concentration of near-optimal selections and a more favorable empirical worst case, suggesting greater robustness across the target datasets.

\begin{table}[ht]
\centering
{\small
\begin{tabular*}{\linewidth}{@{\extracolsep{\fill}}lccc}
\toprule
Method & Selected $\tau_w$ $\uparrow$ & Spearman $\uparrow$ & $p$-value \\
\midrule
\textbf{MetaRank} & \textbf{0.7062} & \textbf{0.5183} & -- \\
SA+SFDA & 0.6426 & -- & 0.0239 \\
ALORS & 0.6493 & 0.4906 & 0.0239 \\
ISAC & 0.6219 & 0.3980 & 0.0239 \\
NCTI & 0.6458 & -- & 0.0482 \\
DISCO & 0.6075 & -- & 0.0491 \\
GB & 0.6100 & 0.4246 & 0.0239 \\
Random & 0.5068 & 0.0416 & 0.0239 \\
\bottomrule
\end{tabular*}
}
\caption{Average metric-selection performance for representative methods. MetaRank achieves the highest selected $\tau_w$ and Spearman correlation, with statistically significant gains under paired Wilcoxon tests with Holm correction. A dash denotes no complete metric ranking.}
\label{tab:main}
\end{table}

No single fixed MTE metric performs consistently well across target datasets, underscoring the importance of task-aware metric selection. Although SA+SFDA achieves the strongest average performance among the fixed-metric baselines, it still exhibits a clear selection quality gap relative to MetaRank. As shown in Fig.~\ref{fig:main-box}, its median selected-metric rank is 3, compared with 2 for MetaRank. Moreover, it never selects the ground-truth best metric on any target dataset. NCTI and DISCO are also competitive on average but remain clearly behind MetaRank. NCTI is concentrated around ranks 4--5, placing most selections away from the top, while its outliers at rank 8 indicate occasional substantial degradation. DISCO has an IQR of 2--6 and reaches rank 12 at the upper whisker, showing even greater variation and occasional selections among the lowest-ranked metrics. Moreover, all fixed metrics achieve lower selected $\tau_w$ than MetaRank's 0.7062 in Table~\ref{tab:main}. Most other fixed metrics perform noticeably worse; full results are provided in the Supplementary Material. Overall, the effectiveness of MTE metrics varies substantially with the target dataset, limiting the reliability of any single fixed metric.

Meta-learning approaches generally outperform most fixed metrics, further supporting the benefit of conditioning metric selection on the target dataset. Among them, ALORS and ISAC are the strongest competing meta-learners, achieving average selected ranks of 3.71 and 4.14, respectively, but both remain behind MetaRank. Their advantage over fixed metrics reflects their ability to leverage experience across meta-training datasets to learn recurring relationships between dataset characteristics and metric utility, allowing metric selection to adapt to an unseen dataset. However, their predictions rely mainly on dataset-level relationships and historical utilities, without directly capturing how each candidate metric ranks the source models on the target dataset. This coarse evidence may limit discrimination among metrics with similar historical performance. MetaRank addresses this limitation with pairwise-order representations that provide behavioral evidence for fine-grained metric selection.

\subsection{Ablation Study and Analysis}
\subsubsection{Effect of Score Representation}
To validate the effectiveness of the pairwise-order representation, we compare MetaRank against two score-level variants: \textbf{Raw Score}, which directly uses original metric values, and \textbf{Standardized Score}, which applies z-score normalization to each score vector $\mathbf{s}_{d,q}$. As shown in Table~\ref{tab:score-representation}, standardizing raw scores improves all three reported measures, suggesting that metric-specific score scales hinder transfer across heterogeneous MTE metrics. Pairwise Order achieves the best overall performance, reducing the average rank of the selected metric from 2.93 to 2.07 while further improving selected $\tau_w$ and Spearman correlation. This demonstrates the benefit of representing metrics through their induced source-model orderings rather than absolute score magnitudes.
\begin{table}[t]
\centering
{\small
\begin{tabular*}{\linewidth}{@{\extracolsep{\fill}}lccc}
\toprule
Representation
& Rank $\downarrow$
& Selected $\tau_w$ $\uparrow$
& Spearman $\uparrow$ \\
\midrule
Raw Score               & 4.07          & 0.6451          & 0.4548 \\
Standardized Score      & 2.93          & 0.6919          & 0.4842 \\
\textbf{Pairwise Order} & \textbf{2.07} & \textbf{0.7062} & \textbf{0.5183} \\
\bottomrule
\end{tabular*}
}
\caption{Effect of score representation. Pairwise Order performs best by encoding source-model orderings rather than score magnitudes.}
\label{tab:score-representation}
\end{table}

\subsubsection{Generalization to Unseen Metrics} To assess MetaRank’s generalization to MTE metrics, we adopt a leave-one-metric-out protocol nested within LODO. In each dataset fold, one metric is held out, and its score vectors and utility labels on meta-training datasets are excluded from training. Since retrieval requires utilities for every metric, we bypass retrieval and rerank the metric pool. During test, all metrics, including the held-out metric, are evaluated on the unseen target dataset and ranked jointly. Although the held-out metric has no historical training examples, MetaRank predicts its utility by matching its target-dataset ordering behavior to those induced by observed metrics on similar meta-training datasets. The metric is selected from the complete metric pool according to predicted utilities. Under this stringent protocol, MetaRank achieves an average selected rank of 3.14 and a selected $\tau_w$ of 0.6908, outperforming the strongest baseline, SA+SFDA, which obtains 3.50 and 0.6426, respectively. These results indicate that MetaRank effectively transfers learned knowledge to unseen metrics without retraining.

\subsubsection{Retrieve-and-Rerank Cascade}
To investigate the roles of retrieval and reranking, we compare MetaRank with three variants: (i) \textbf{Retrieve-Only}, which selects the top retrieved metric based solely on dataset similarity; (ii) \textbf{Random-Retrieve}, which reranks five randomly sampled metrics; and (iii) \textbf{Rerank-Only}, which reranks all 13 metrics. Table~\ref{tab:cascade} shows that MetaRank substantially outperforms Retrieve-Only, confirming the benefit of fine-grained reranking. Its advantage over Random-Retrieve under the same budget further demonstrates the importance of similarity-guided candidate filtering. MetaRank also achieves a lower selected rank and higher selected $\tau_w$ than Rerank-Only while evaluating only 5 instead of 13 metrics, despite its lower global Spearman correlation. This favorable trade-off aligns with our primary objective of selecting the top-performing metric rather than recovering the complete metric ranking. The results suggest that retrieval sharpens reranking by screening out weak candidates, thereby reducing both ranking noise and computation. Additional retrieval-budget analysis is provided in the Supplementary Material.

\begin{table}[t]
\centering
{\small
\setlength{\tabcolsep}{3.5pt}
\begin{tabular*}{\linewidth}{@{\extracolsep{\fill}}lcccc@{}}
\toprule
Setting
& Budget
& Rank $\downarrow$
& Selected $\tau_w$ $\uparrow$
& Spearman $\uparrow$ \\
\midrule
Retrieve-Only     & 1  & 5.00          & 0.6065          & 0.4514 \\
Random-Retrieve   & 5  & 3.66          & 0.6581          & 0.2502 \\
Rerank-Only       & 13 & 2.64          & 0.7014          & \textbf{0.7691} \\
\textbf{MetaRank} & 5  & \textbf{2.07} & \textbf{0.7062} & 0.5183 \\
\bottomrule
\end{tabular*}
}
\caption{Ablation of the retrieve-and-rerank cascade. Rerank-Only better recovers the full metric ranking, whereas MetaRank more reliably identifies the top metric.}
\label{tab:cascade}
\end{table}

\subsubsection{Sensitivity to the Ridge Parameter}
We set the KRR ridge parameter to $\lambda=1/L$, where $L$ is the number of meta-training datasets. This scaling reflects that datasets, rather than the $LN$ correlated dataset--metric records, constitute the independent meta-learning units, while keeping the regularization strength invariant to the metric-pool size. To assess sensitivity, we vary $\lambda$ over $\{0.5/L,1/L,2/L\}$. All three settings yield the same average selected-metric rank of 2.07, selected $\tau_w$ of 0.7062, and Spearman correlation of 0.5183, indicating that MetaRank is not sensitive to moderate changes in ridge regularization. Additional sensitivity analyses of the two kernel bandwidths, $h_c$ and $h_\pi$, are provided in the Supplementary Material.

\subsubsection{Self-Supervised Model Zoo}
To assess MetaRank on self-supervised learning (SSL) models, we train and evaluate it in a setup comprising 8 SSL pre-trained source models, 10 target datasets, and 9 applicable MTE metrics. MetaRank achieves a superior average metric rank of 1.20, outperforming the top fixed metric DISCO (1.30) and the leading meta-learner ALORS (1.50). Furthermore, MetaRank identifies the ground-truth best metric on 8 of 10 target datasets and selects a Top-3 metric in all cases. These results demonstrate its applicability to a self-supervised model zoo. 

\section{Conclusion}
In this paper, we observe that the effectiveness of model transferability estimation (MTE) metrics is highly task-dependent, and no single metric is universally optimal across target datasets. We therefore propose MetaRank, a task-aware framework for selecting MTE metrics through a retrieve-and-rerank cascade. MetaRank first retrieves promising metrics by transferring historical utility patterns from similar datasets, and then reranks them according to the source-model orderings they induce on the target dataset. Its pairwise-order representation places heterogeneous metrics in a shared behavioral space, enabling utility evidence to be transferred across metrics exhibiting comparable ordering behavior. Extensive experiments and analyses demonstrate the effectiveness and generalization of MetaRank.

\bibliography{aaai2027}

@article{zhuang2020comprehensive,
  title={A comprehensive survey on transfer learning},
  author={Zhuang, Fuzhen and Qi, Zhiyuan and Duan, Keyu and Xi, Dongbo and Zhu, Yongchun and Zhu, Hengshu and Xiong, Hui and He, Qing},
  journal={Proceedings of the IEEE},
  volume={109},
  number={1},
  pages={43--76},
  year={2020},
  publisher={IEEE}
}

@article{pan2009survey,
  author={Pan, Sinno Jialin and Yang, Qiang},
  journal={IEEE Transactions on Knowledge and Data Engineering}, 
  title={A Survey on Transfer Learning}, 
  year={2010},
  volume={22},
  number={10},
  pages={1345--1359},
  doi={10.1109/TKDE.2009.191}}

@ARTICLE{freelunch,
  author={Wolpert, D.H. and Macready, W.G.},
  journal={IEEE Transactions on Evolutionary Computation}, 
  title={No free lunch theorems for optimization}, 
  year={1997},
  volume={1},
  number={1},
  pages={67--82},
  doi={10.1109/4235.585893}}

@inproceedings{shao2022SFDA,
  title={Not all models are equal: Predicting model transferability in a self-challenging Fisher space},
  author={Shao, Wenqi and Zhao, Xun and Ge, Yixiao and Zhang, Zhaoyang and Yang, Lei and Wang, Xiaogang and Shan, Ying and Luo, Ping},
  booktitle={European Conference on Computer Vision},
  pages={286--302},
  year={2022},
  organization={Springer}
}

@inproceedings{tran2019NCE,
  title={Transferability and hardness of supervised classification tasks},
  author={Tran, Anh T and Nguyen, Cuong V and Hassner, Tal},
  booktitle={Proceedings of the IEEE/CVF International Conference on Computer Vision},
  pages={1395--1405},
  year={2019}
}

@inproceedings{pandy2022GBC,
  title={Transferability estimation using Bhattacharyya class separability},
  author={P{\'a}ndy, Michal and Agostinelli, Andrea and Uijlings, Jasper and Ferrari, Vittorio and Mensink, Thomas},
  booktitle={Proceedings of the IEEE/CVF Conference on Computer Vision and Pattern Recognition},
  pages={9172--9182},
  year={2022}
}

@inproceedings{bao2019HScore,
  title={An information-theoretic approach to transferability in task transfer learning},
  author={Bao, Yajie and Li, Yang and Huang, Shao-Lun and Zhang, Lin and Zheng, Lizhong and Zamir, Amir and Guibas, Leonidas},
  booktitle={2019 IEEE International Conference on Image Processing (ICIP)},
  pages={2309--2313},
  year={2019},
  organization={IEEE}
}

@inproceedings{wang2023NCTI,
  title={How far pre-trained models are from neural collapse on the target dataset informs their transferability},
  author={Wang, Zijian and Luo, Yadan and Zheng, Liang and Huang, Zi and Baktashmotlagh, Mahsa},
  booktitle={Proceedings of the IEEE/CVF International Conference on Computer Vision},
  pages={5549--5558},
  year={2023}
}

@inproceedings{nguyen2020leep,
  title={LEEP: A new measure to evaluate transferability of learned representations},
  author={Nguyen, Cuong and Hassner, Tal and Seeger, Matthias and Archambeau, Cedric},
  booktitle={International Conference on Machine Learning},
  pages={7294--7305},
  year={2020},
  organization={PMLR}
}

@inproceedings{li2021nleep,
  title={Ranking neural checkpoints},
  author={Li, Yandong and Jia, Xuhui and Sang, Ruoxin and Zhu, Yukun and Green, Bradley and Wang, Liqiang and Gong, Boqing},
  booktitle={Proceedings of the IEEE/CVF Conference on Computer Vision and Pattern Recognition},
  pages={2663--2673},
  year={2021}
}

@inproceedings{you2021logme,
  title={LogME: Practical assessment of pre-trained models for transfer learning},
  author={You, Kaichao and Liu, Yong and Wang, Jianmin and Long, Mingsheng},
  booktitle={International Conference on Machine Learning},
  pages={12133--12143},
  year={2021},
  organization={PMLR}
}

@inproceedings{gholami2023etran,
  title={ETran: Energy-based transferability estimation},
  author={Gholami, Mohsen and Akbari, Mohammad and Wang, Xinglu and Kamranian, Behnam and Zhang, Yong},
  booktitle={Proceedings of the IEEE/CVF International Conference on Computer Vision},
  pages={18613--18622},
  year={2023}
}

@misc{guo2024kite,
      title={KITE: A Kernel-based Improved Transferability Estimation Method}, 
      author={Yunhui Guo},
      year={2024},
      eprint={2405.01603},
      archivePrefix={arXiv},
      primaryClass={cs.LG},
      url={https://arxiv.org/abs/2405.01603}, 
}

@article{zhang2023modelspider,
  title={Model Spider: Learning to rank pre-trained models efficiently},
  author={Zhang, Yi-Kai and Huang, Ting-Ji and Ding, Yao-Xiang and Zhan, De-Chuan and Ye, Han-Jia},
  journal={Advances in Neural Information Processing Systems},
  volume={36},
  pages={13692--13719},
  year={2023}
}

@article{ding2022synlearn,
  title={Pre-trained model reusability evaluation for small-data transfer learning},
  author={Ding, Yao-Xiang and Wu, Xi-Zhu and Zhou, Kun and Zhou, Zhi-Hua},
  journal={Advances in Neural Information Processing Systems},
  volume={35},
  pages={37389--37400},
  year={2022}
}

@inproceedings{vigna2015kendall,
  title={A weighted correlation index for rankings with ties},
  author={Vigna, Sebastiano},
  booktitle={Proceedings of the 24th International Conference on World Wide Web},
  pages={1166--1176},
  year={2015}
}

@misc{ding2024model,
      title={Which Model to Transfer? A Survey on Transferability Estimation}, 
      author={Yuhe Ding and Bo Jiang and Aijing Yu and Aihua Zheng and Jian Liang},
      year={2024},
      eprint={2402.15231},
      archivePrefix={arXiv},
      primaryClass={cs.LG},
      url={https://arxiv.org/abs/2402.15231}, 
}

@misc{maji2013aircraft,
      title={Fine-Grained Visual Classification of Aircraft}, 
      author={Subhransu Maji and Esa Rahtu and Juho Kannala and Matthew Blaschko and Andrea Vedaldi},
      year={2013},
      eprint={1306.5151},
      archivePrefix={arXiv},
      primaryClass={cs.CV},
      url={https://arxiv.org/abs/1306.5151}, 
}

@article{krause2013cars,
  title={Collecting a large-scale dataset of fine-grained cars},
  author={Krause, Jonathan and Deng, Jia and Stark, Michael and Fei-Fei, Li},
  year={2013}
}

@inproceedings{fei2004caltech101,
  title={Learning generative visual models from few training examples: An incremental Bayesian approach tested on 101 object categories},
  author={Fei-Fei, Li and Fergus, Rob and Perona, Pietro},
  booktitle={2004 Conference on Computer Vision and Pattern Recognition Workshop},
  pages={178--178},
  year={2004},
  organization={IEEE}
}

@article{krizhevsky2009cifar,
  title={Learning multiple layers of features from tiny images},
  author={Krizhevsky, Alex and Hinton, Geoffrey and others},
  year={2009},
  publisher={Toronto, ON, Canada}
}

@inproceedings{cimpoi2014dtd,
  title={Describing textures in the wild},
  author={Cimpoi, Mircea and Maji, Subhransu and Kokkinos, Iasonas and Mohamed, Sammy and Vedaldi, Andrea},
  booktitle={Proceedings of the IEEE Conference on Computer Vision and Pattern Recognition},
  pages={3606--3613},
  year={2014}
}

@inproceedings{nilsback2008flower,
  author={Nilsback, Maria-Elena and Zisserman, Andrew},
  booktitle={2008 Sixth Indian Conference on Computer Vision, Graphics \& Image Processing}, 
  title={Automated Flower Classification over a Large Number of Classes}, 
  year={2008},
  pages={722--729},
  doi={10.1109/ICVGIP.2008.47}}

@inproceedings{bossard2014food,
  title={Food-101--mining discriminative components with random forests},
  author={Bossard, Lukas and Guillaumin, Matthieu and Van Gool, Luc},
  booktitle={European Conference on Computer Vision},
  pages={446--461},
  year={2014},
  organization={Springer}
}

@inproceedings{parkhi2012pets,
  title={Cats and dogs},
  author={Parkhi, Omkar M and Vedaldi, Andrea and Zisserman, Andrew and Jawahar, CV},
  booktitle={2012 IEEE Conference on Computer Vision and Pattern Recognition},
  pages={3498--3505},
  year={2012},
  organization={IEEE}
}

@inproceedings{xiao2010sun,
  author={Xiao, Jianxiong and Hays, James and Ehinger, Krista A. and Oliva, Aude and Torralba, Antonio},
  booktitle={2010 IEEE Computer Society Conference on Computer Vision and Pattern Recognition}, 
  title={SUN database: Large-scale scene recognition from abbey to zoo}, 
  year={2010},
  pages={3485--3492},
  doi={10.1109/CVPR.2010.5539970}}

@article{everingham2010voc,
  title={The PASCAL visual object classes (VOC) challenge},
  author={Everingham, Mark and Van Gool, Luc and Williams, Christopher KI and Winn, John and Zisserman, Andrew},
  journal={International Journal of Computer Vision},
  volume={88},
  number={2},
  pages={303--338},
  year={2010},
  publisher={Springer}
}

@inproceedings{he2016res,
  title={Deep residual learning for image recognition},
  author={He, Kaiming and Zhang, Xiangyu and Ren, Shaoqing and Sun, Jian},
  booktitle={Proceedings of the IEEE Conference on Computer Vision and Pattern Recognition},
  pages={770--778},
  year={2016}
}

@inproceedings{huang2017densely,
  title={Densely connected convolutional networks},
  author={Huang, Gao and Liu, Zhuang and van der Maaten, Laurens and Weinberger, Kilian Q},
  booktitle={Proceedings of the IEEE Conference on Computer Vision and Pattern Recognition},
  pages={4700--4708},
  year={2017}
}

@inproceedings{tan2019mnasnet,
  title={MnasNet: Platform-aware neural architecture search for mobile},
  author={Tan, Mingxing and Chen, Bo and Pang, Ruoming and Vasudevan, Vijay and Sandler, Mark and Howard, Andrew and Le, Quoc V},
  booktitle={Proceedings of the IEEE/CVF Conference on Computer Vision and Pattern Recognition},
  pages={2820--2828},
  year={2019}
}

@inproceedings{sandler2018mobilenetv2,
  title={MobileNetV2: Inverted residuals and linear bottlenecks},
  author={Sandler, Mark and Howard, Andrew and Zhu, Menglong and Zhmoginov, Andrey and Chen, Liang-Chieh},
  booktitle={Proceedings of the IEEE Conference on Computer Vision and Pattern Recognition},
  pages={4510--4520},
  year={2018}
}

@inproceedings{szegedy2015google,
  title={Going deeper with convolutions},
  author={Szegedy, Christian and Liu, Wei and Jia, Yangqing and Sermanet, Pierre and Reed, Scott and Anguelov, Dragomir and Erhan, Dumitru and Vanhoucke, Vincent and Rabinovich, Andrew},
  booktitle={Proceedings of the IEEE Conference on Computer Vision and Pattern Recognition},
  pages={1--9},
  year={2015}
}

@inproceedings{szegedy2016inception,
  title={Rethinking the inception architecture for computer vision},
  author={Szegedy, Christian and Vanhoucke, Vincent and Ioffe, Sergey and Shlens, Jon and Wojna, Zbigniew},
  booktitle={Proceedings of the IEEE Conference on Computer Vision and Pattern Recognition},
  pages={2818--2826},
  year={2016}
}

@article{abou2024onesize,
  title={One size does not fit all in evaluating model selection scores for image classification},
  author={Abou Baker, Nermeen and Handmann, Uwe},
  journal={Scientific Reports},
  volume={14},
  number={1},
  pages={30239},
  year={2024},
  publisher={Nature Publishing Group UK London}
}

@article{hospedales2021metasurvey,
  author={Hospedales, Timothy and Antoniou, Antreas and Micaelli, Paul and Storkey, Amos},
  journal={IEEE Transactions on Pattern Analysis and Machine Intelligence}, 
  title={Meta-Learning in Neural Networks: A Survey}, 
  year={2022},
  volume={44},
  number={9},
  pages={5149--5169},
  doi={10.1109/TPAMI.2021.3079209}
  }

@inproceedings{achille2019task2vec,
  title={Task2Vec: Task embedding for meta-learning},
  author={Achille, Alessandro and Lam, Michael and Tewari, Rahul and Ravichandran, Avinash and Maji, Subhransu and Fowlkes, Charless C and Soatto, Stefano and Perona, Pietro},
  booktitle={Proceedings of the IEEE/CVF International Conference on Computer Vision},
  pages={6430--6439},
  year={2019}
}

@inproceedings{qin2024metaood,
title={Meta{OOD}: Automatic Selection of {OOD} Detection Models},
author={Yuehan Qin and Yichi Zhang and Yi Nian and Xueying Ding and Yue Zhao},
booktitle={International Conference on Learning Representations},
year={2025},
url={https://openreview.net/forum?id=9qpdDiDQ2H}
}

@article{vettoruzzo2024advances,
  title={Advances and challenges in meta-learning: A technical review},
  author={Vettoruzzo, Anna and Bouguelia, Mohamed-Rafik and Vanschoren, Joaquin and R{\"o}gnvaldsson, Thorsteinn and Santosh, KC},
  journal={IEEE Transactions on Pattern Analysis and Machine Intelligence},
  volume={46},
  number={7},
  pages={4763--4779},
  year={2024},
  publisher={IEEE}
}

@incollection{kadioglu2010isac,
  title={ISAC--instance-specific algorithm configuration},
  author={Kadioglu, Serdar and Malitsky, Yuri and Sellmann, Meinolf and Tierney, Kevin},
  booktitle={ECAI 2010},
  pages={751--756},
  year={2010},
  publisher={IOS Press}
}

@article{nikolic2013as,
  title={Simple algorithm portfolio for SAT},
  author={Nikoli{\'c}, Mladen and Mari{\'c}, Filip and Jani{\v{c}}i{\'c}, Predrag},
  journal={Artificial Intelligence Review},
  volume={40},
  number={4},
  pages={457--465},
  year={2013},
  publisher={Springer}
}

@article{misir2017alors,
  title={ALORS: An algorithm recommender system},
  author={M{\i}s{\i}r, Mustafa and Sebag, Mich{\`e}le},
  journal={Artificial Intelligence},
  volume={244},
  pages={291--314},
  year={2017},
  publisher={Elsevier}
}

@inproceedings{he2017ncf,
  title={Neural collaborative filtering},
  author={He, Xiangnan and Liao, Lizi and Zhang, Hanwang and Nie, Liqiang and Hu, Xia and Chua, Tat-Seng},
  booktitle={Proceedings of the 26th International Conference on World Wide Web},
  pages={173--182},
  year={2017}
}

@article{jomaa2021dataset2vec,
  title={Dataset2Vec: Learning dataset meta-features},
  author={Jomaa, Hadi S and Schmidt-Thieme, Lars and Grabocka, Josif},
  journal={Data Mining and Knowledge Discovery},
  volume={35},
  number={3},
  pages={964--985},
  year={2021},
  publisher={Springer}
}

@inproceedings{kornblith2019better,
  title={Do better ImageNet models transfer better?},
  author={Kornblith, Simon and Shlens, Jonathon and Le, Quoc V},
  booktitle={Proceedings of the IEEE/CVF Conference on Computer Vision and Pattern Recognition},
  pages={2661--2671},
  year={2019}
}

@inproceedings{hu2019strategies,
title={Strategies for Pre-training Graph Neural Networks},
author={Hu, Weihua and Liu, Bowen and Gomes, Joseph and Zitnik, Marinka and Liang, Percy and Pande, Vijay and Leskovec, Jure},
booktitle={International Conference on Learning Representations},
year={2020},
url={https://openreview.net/forum?id=HJlWWJSFDH},
}

@inproceedings{zhang2025disco,
  title={Assessing pre-trained models for transfer learning through distribution of spectral components},
  author={Zhang, Tengxue and Shu, Yang and Chen, Xinyang and Long, Yifei and Guo, Chenjuan and Yang, Bin},
  booktitle={Proceedings of the AAAI Conference on Artificial Intelligence},
  number={21},
  pages={22560--22568},
  year={2025},
  doi={10.1609/aaai.v39i21.34414}
}

@inproceedings{ibrahim2022newer,
  title={Newer is not always better: Rethinking transferability metrics, their peculiarities, stability and performance},
  author={Ibrahim, Shibal and Ponomareva, Natalia and Mazumder, Rahul},
  booktitle={Joint European Conference on Machine Learning and Knowledge Discovery in Databases},
  pages={693--709},
  year={2022},
  organization={Springer}
}

@inproceedings{khoba2025SA,
  title={Feature space perturbation: a panacea to enhanced transferability estimation},
  author={Khoba, Prafful Kumar and Wang, Zijian and Arora, Chetan and Baktashmotlagh, Mahsa},
  booktitle={2025 IEEE/CVF Winter Conference on Applications of Computer Vision (WACV)},
  pages={1299--1308},
  year={2025},
  organization={IEEE}
}

@article{helber2019eurosat,
  author={Helber, Patrick and Bischke, Benjamin and Dengel, Andreas and Borth, Damian},
  journal={IEEE Journal of Selected Topics in Applied Earth Observations and Remote Sensing}, 
  title={EuroSAT: A Novel Dataset and Deep Learning Benchmark for Land Use and Land Cover Classification}, 
  year={2019},
  volume={12},
  number={7},
  pages={2217--2226},
  doi={10.1109/JSTARS.2019.2918242}}

@inproceedings{coates2011stl10,
  title={An analysis of single-layer networks in unsupervised feature learning},
  author={Coates, Adam and Ng, Andrew and Lee, Honglak},
  booktitle={Proceedings of the Fourteenth International Conference on Artificial Intelligence and Statistics},
  pages={215--223},
  year={2011},
  organization={JMLR Workshop and Conference Proceedings}
}

@misc{gutman2016ISIC,
      title={Skin Lesion Analysis toward Melanoma Detection: A Challenge at the International Symposium on Biomedical Imaging (ISBI) 2016, hosted by the International Skin Imaging Collaboration (ISIC)}, 
      author={David Gutman and Noel C. F. Codella and Emre Celebi and Brian Helba and Michael Marchetti and Nabin Mishra and Allan Halpern},
      year={2016},
      eprint={1605.01397},
      archivePrefix={arXiv},
      primaryClass={cs.CV},
      url={https://arxiv.org/abs/1605.01397}, 
}

@InProceedings{transrate,
  title = 	 {Frustratingly Easy Transferability Estimation},
  author =       {Huang, Long-Kai and Huang, Junzhou and Rong, Yu and Yang, Qiang and Wei, Ying},
  booktitle = 	 {International Conference on Machine Learning},
  pages = 	 {9201--9225},
  year = 	 {2022},
  editor = 	 {Chaudhuri, Kamalika and Jegelka, Stefanie and Song, Le and Szepesvari, Csaba and Niu, Gang and Sabato, Sivan},
  volume = 	 {162},
  series = 	 {Proceedings of Machine Learning Research},
  month = 	 {17--23 Jul},
  publisher =    {PMLR},
  url = 	 {https://proceedings.mlr.press/v162/huang22d.html}
}

@inproceedings{parc,
 author = {Bolya, Daniel and Mittapalli, Rohit and Hoffman, Judy},
 booktitle = {Advances in Neural Information Processing Systems},
 editor = {M. Ranzato and A. Beygelzimer and Y. Dauphin and P.S. Liang and J. Wortman Vaughan},
 pages = {19301--19312},
 publisher = {Curran Associates, Inc.},
 title = {Scalable Diverse Model Selection for Accessible Transfer Learning},
 url = {https://proceedings.neurips.cc/paper_files/paper/2021/file/a1140a3d0df1c81e24ae954d935e8926-Paper.pdf},
 volume = {34},
 year = {2021}
}

@inproceedings{clip,
  title={Learning transferable visual models from natural language supervision},
  author={Radford, Alec and Kim, Jong Wook and Hallacy, Chris and Ramesh, Aditya and Goh, Gabriel and Agarwal, Sandhini and Sastry, Girish and Askell, Amanda and Mishkin, Pamela and Clark, Jack and others},
  booktitle={International Conference on Machine Learning},
  pages={8748--8763},
  year={2021},
  organization={PMLR}
}


\end{document}